\documentclass[letterpaper, 10 pt, conference]{ieeeconf} 

\makeatletter
\let\NAT@parse\undefined
\makeatother

\IEEEoverridecommandlockouts 
\usepackage{times}

\usepackage[numbers,sort&compress]{natbib}
\usepackage{multicol}
\usepackage{url}

\usepackage{tikz}
\usepackage{siunitx}
\usepackage{svg}
\usepackage{multirow}
\usepackage{subfig}
\usepackage{placeins}
\usepackage{pgfplots}
\usepackage{hyperref}
\hypersetup{bookmarks=false}

\usepackage[acronym]{glossaries}
\newacronym{rl}{RL}{Reinforcement Learning}
\newacronym{mpc}{MPC}{Model Predictive Control}
\newacronym{nmpc}{NMPC}{Nonlinear Model Predictive Control}
\newacronym{dmpc}{DMPC}{Deep Value Model Predictive Control}
\newacronym{mav}{MAV}{Micro Aerial Vehicle}
\newacronym{slq}{SLQ}{Sequential Linear Quadratic}
\newacronym{ram}{RAM}{Random Access Memory}
\newacronym{wbc}{WBC}{Whole-Body Control}
\newacronym{ros}{ROS}{Robot Operating System}
\newacronym{slam}{SLAM}{Simultaneous Localization and Mapping}
\newacronym{tsdf}{TSDF}{Truncated Signed Distance Field}
\newacronym{rms}{RMS}{Root Mean Square}
\newacronym{mdp}{MDP}{Markov Decision Process}
\newacronym{arc}{ARC}{Amazon Robotics Challenge}
\newacronym{dof}{DoF}{Degrees of Freedom}
\newacronym{ppo}{PPO}{Proximal Policy Optimization}
\newacronym{ddpg}{DDPG}{Deep Deterministic Policy Gradient}
\newacronym{trpo}{TRPO}{Trust Region Policy Optimization}
\newacronym{ac}{AC}{Actor-Critic}
\newacronym{mlp}{MLP}{Multi-layer perceptron}
\newacronym{rnn}{RNN}{Recurrent Neural Network}
\newacronym{adr}{ADR}{Automatic Domain Randomization}
\newacronym{hpt}{HPT}{Harmonic Potential Field}
\newacronym{lidar}{LiDAR}{Light Detection And Ranging}

\newcommand{\expnumber}[2]{{#1}\mathrm{e}{#2}}
\newcommand{\norm}[1]{\left\lVert{#1}\right\rVert}
\newcommand{\abs}[1]{\left|{#1}\right|}
\newcommand{\abssmall}[1]{|{#1}|}
\input{pgfplots_preamble.tex}
\pgfplotsset{every tick label/.append style={font=\footnotesize}}

\begin{document}

\title{\LARGE \bf
Whole-Body Control of a Mobile Manipulator using End-to-End Reinforcement Learning}

\author{Julien Kindle, Fadri Furrer, Tonci Novkovic, Jen Jen Chung, Roland Siegwart and Juan Nieto
\thanks{This work was supported by ABB Corporate Research.}
\thanks{The authors are with the Autonomous Systems Lab, ETH Z{\" u}rich, Z{\"u}rich 8092, Switzerland. {\tt\small\{jkindle; fadri; ntonci; chungj; rsiegwart; jnieto\}@ethz.ch}}%
}

\maketitle

\begin{abstract}
Mobile manipulation is usually achieved by sequentially executing base and manipulator movements. This simplification, however, leads to a loss in efficiency and in some cases a reduction of workspace size. Even though different methods have been proposed to solve \acrfull{wbc} online, they are either limited by a kinematic model or do not allow for reactive, online obstacle avoidance. In order to overcome these drawbacks, in this work, we propose an end-to-end \acrfull{rl} approach to \acrshort{wbc}. We compared our learned controller against a state-of-the-art sampling-based method in simulation and achieved faster overall mission times. In addition, we validated the learned policy on our mobile manipulator RoyalPanda in challenging narrow corridor environments.

\end{abstract}

\section{Introduction}
\label{sec:intro}

According to the International Federation of Robotics, 16.3M service robots for personal and domestic use were sold in 2018, which is a 59\% increase compared to the previous year~\cite{IFR2019}. The majority of these commercially deployed robots target applications with a focus on mobility such as autonomous vacuum cleaner robots~\cite{irobot} and industrial transportation systems~\cite{dandrea2012kiva}. However, there exist far fewer mobile robots that incorporate advanced manipulation capabilities.
Several research platforms~\cite{bohren2011towards, wise2016fetch} have been developed for mobile manipulation tasks such as opening doors~\cite{meeussen2010autonomous}. These typically decouple the movement of the base from the movement of the manipulator to simplify the control problem. However, in doing so, these approaches miss out on any potential synergies that arise from controlling the entire robot as a whole. Indeed, \acrfull{wbc} aims at improving the efficiency and performance of such mobile manipulation robots by planning for and controlling all \acrfull{dof} of the robot simultaneously.


There are several classes of approaches to whole-body trajectory planning and control. Sampling-based methods~\cite{Kuffner2000RRTC, LaValle1998RapidlyexploringRT, kavraki1996PRM, jaillet2010confspacecostmap} grow a tree of collision-free joint configurations. 
However, these methods not only suffer from the curse of dimensionality but also require full knowledge of the environment at the planning stage. In order to operate in an unknown and possibly dynamic environment, which is observed only via sensors mounted to the robot, offline trajectory planning is often insufficient.

Another approach to tackle \acrshort{wbc} is \acrfull{mpc}. In this field, the target is to solve the planning problem online by finding the sequence of controls that optimize a given objective function, e.g. minimizing deviation from a target trajectory. Even though multiple \acrshort{mpc} formulations have been proposed for solving \acrshort{wbc} online, these are either limited in their obstacle avoidance capabilities~\cite{Farshidian2017RealtimeMP} or only used a kinematic model~\cite{avanzini2015constraint}.

Recent works have shown that learning-based methods can compete against classical algorithms for local path planning and manipulation tasks~\cite{tai2017virtual,pfeiffer2017perception,iriondo2019pickandplace,novkovic2019object,breyer2019comparing,popov2017data,gu2017deep}. In particular, \acrfull{rl} has shown potential in fields where data collection and labelling for supervised learning is expensive.

\begin{figure}[t]
    \centering
    \includegraphics[width=0.9\columnwidth]{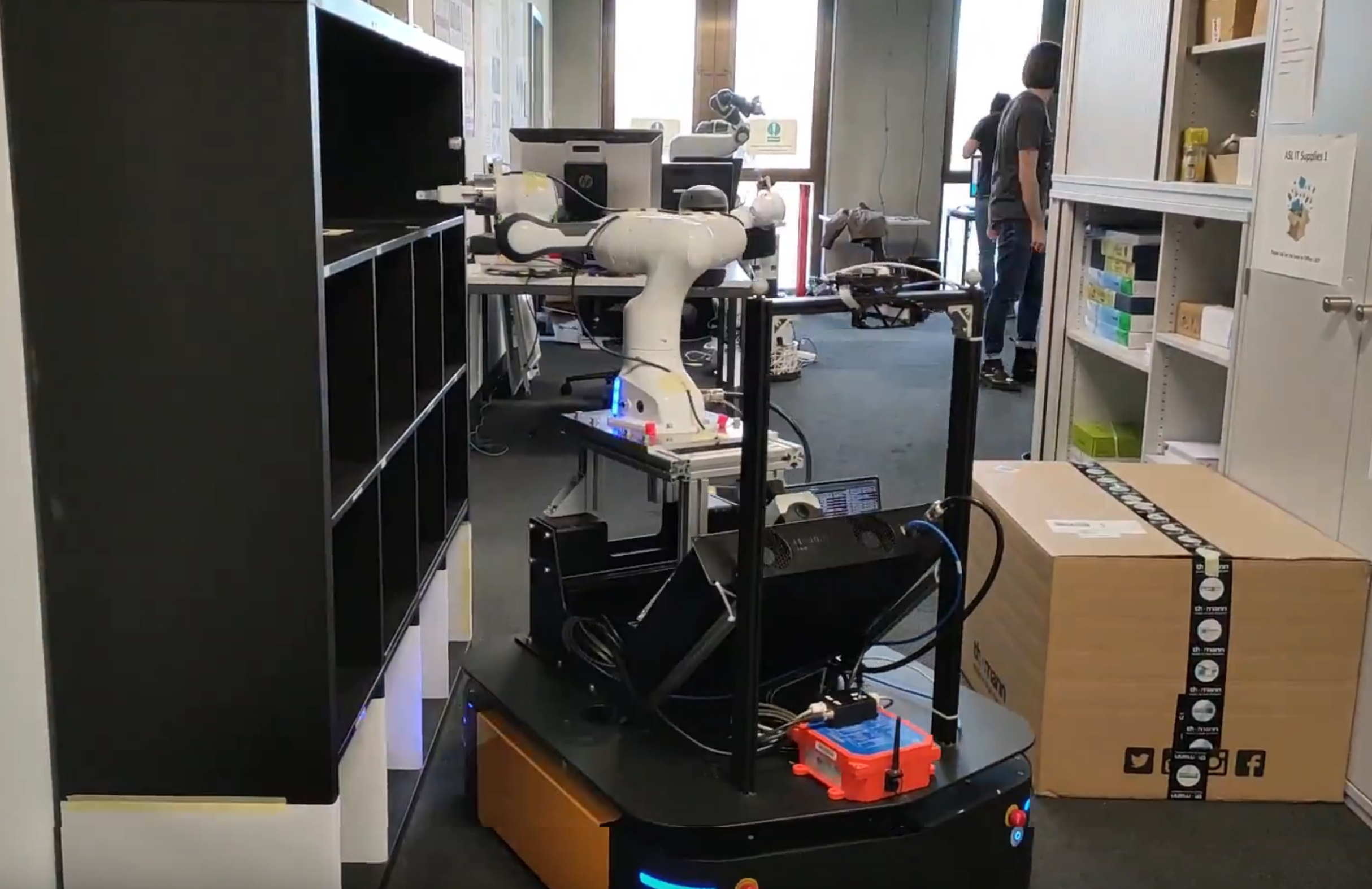}
    \caption{RoyalPanda executing a learned \acrshort{wbc} policy for reaching a setpoint in the shelf.}
    \label{fig:teaser}
\end{figure}

In this work, we hypothesize that learning-based methods may be viewed as an alternative to sampling-based offline trajectory planning and online \acrshort{mpc}. We use \acrshort{rl} to train a controller for a mobile manipulator which observes its surroundings using two \acrshort{lidar} scans and steers all \acrshort{dof} made available to the robot. For that, we created a simulation of our mobile manipulator RoyalPanda (Franka Emika Panda arm mounted on a Clearpath Ridgeback platform) in randomized corridor environments and trained a \acrshort{wbc} policy to successfully steer the robot to commanded end-effector setpoints within it.



We make the following contributions:
\begin{itemize}
    \item We present a learning-based approach to \acrshort{wbc} that runs online at a rate of at least 100 Hz on an embedded device (Raspberry Pi 3B).
    \item We demonstrate the faster overall mission times (planning plus execution) of our trained controller as compared to a sampling-based method in simulation, and
    \item We show that the learned policy can be directly transferred to a real robot and achieve similar performance as in simulation.
\end{itemize}

\section{Related Work}
A simple yet widely used approach to mobile manipulation is the sequential execution of base and arm movements as in~\cite{quigley2009high, wang2010visualservo, li2017reinforcement}. However, this approach is slow and, depending on the robot and workspace configuration, may require multiple iterations in order to reach a given end-effector goal. 

A way to improve execution time and work space size is \acrshort{wbc}, the simultaneous control of both the base and arm on a mobile manipulator. Classical sampling-based methods such as those offered by MoveIt!~\cite{moveit} can be modified to account for a (holonomic) moving base of the manipulator. \citet{yang2016humanoid} adapted the random sampler together with the sampling space and the interpolation function for common sampling-based algorithms~\cite{Kuffner2000RRTC, LaValle1998RapidlyexploringRT, kavraki1996PRM, sucan1070kinodym, hsu1997ecs, ante2001prmlcc} to generate collision-free and balanced whole-body trajectories for humanoid robots. However, these algorithms generate an offline trajectory and therefore require a static and known environment.

On the other hand, \acrshort{mpc} models the dynamics of a system together with the state boundaries as constraints over a fixed horizon and aims to minimize a cost function which is dependent on the states and inputs of the system. 
Examples in robotics include high-level control for fixed-wing guidance~\cite{stastny2018nonlinear} and multi-\acrshort{mav} collision avoidance and trajectory tracking~\cite{kamel2017nonlinear}. For mobile manipulation, several different approaches have been presented. 
\citet{Minniti2019WholeBodyMF} propose an optimal control framework which allows \acrshort{wbc} for end-effector tasks while simultaneously stabilizing the unstable base of a balancing robot in a single optimization problem. A \acrfull{slq} method described by \citet{Farshidian2017RealtimeMP} was used as a solver for the optimal control problem. However, the final controller does not incorporate obstacles.
\citet{avanzini2015constraint} formulated an \acrshort{mpc} which drives to a reference point while avoiding obstacles. The controller keeps the arm in a safe configuration by dynamically varying a weight matrix of the optimization problem as a function of goal distance. Their model formulation, however, is limited to a kinematic model.

\acrshort{rl} has also been a popular choice for learning manipulation skills as well as navigation policies. \citet{novkovic2019object} proposed an \acrshort{rl} approach to object finding in environments which require physical interaction with the end-effector to expose the searched object. \citet{breyer2019comparing} use \acrshort{rl} with curriculum learning to train a controller from sparse rewards for end-effector object grasping. \citet{popov2017data} used a modified variant of \acrfull{ddpg} to learn a control policy for grasping and stacking objects with a manipulator, while \citet{gu2017deep} developed an approach to train a controller with asynchronous \acrshort{rl} directly on multiple robot arms.

In terms of learning navigation policies, \citet{tai2017virtual} trained a mapless motion planner to drive a differential base to a setpoint given only $10$ distance measurements. \citet{pfeiffer2017perception} generated data from the ROS navigation stack to train a controller in a supervised fashion to steer a differential drive robot from \acrshort{lidar} scans. \citet{iriondo2019pickandplace} trained a controller for the base of a mobile manipulator that drives the platform to a feasible position for grasping an object on a table. However, in their paper, the tests were only conducted in simulation, the absolute position of the robot was given to the controller and the structure of the environment was fixed throughout the training.

Recently, progress has been made in the field of \acrshort{mpc} in combination with machine learning, especially \acrshort{rl}. \citet{farshidian2019deep} developed an algorithm called \acrfull{dmpc} that uses an actor-critic agent whose critic is a neural network modelling a value function and whose actor is an \acrshort{mpc} that uses the value function as part of its optimization cost. The \acrshort{mpc} is then solved by using \acrshort{slq} where the Hessian and Jacobian of the value function network are computed using an automatic differentiation library. In their work, the controller is trained for a single static environment. 

Our approach is based on the work of \citet{pfeiffer2017perception} and \citet{popov2017data}. We combine the information of two \acrshort{lidar} scans, the system's internal states and the setpoint in end-effector frame in a deep neural network. The network is trained in simulation using \acrshort{rl} as a \acrshort{wbc} policy to drive the end-effector to the setpoint while avoiding obstacles.

\section{Method}
\label{sec:method}
In this work, we use \acrshort{rl} to discover \acrshort{wbc} policies that can steer a mobile manipulator to target end-effector poses while avoiding obstacles. To do so, our system uses raw $2$D \acrshort{lidar} and the target position in end-effector frame as input to the \acrshort{rl} algorithm. The agent (Section~\ref{sec:method_agent}) is trained in a simulated randomized corridor environment (Section~\ref{sec:method_simulation}).

\subsection{Observation and action space}
\begin{figure}[b]
  \centering
  \includegraphics[trim=30 230 20 170,clip,width=0.8\columnwidth]{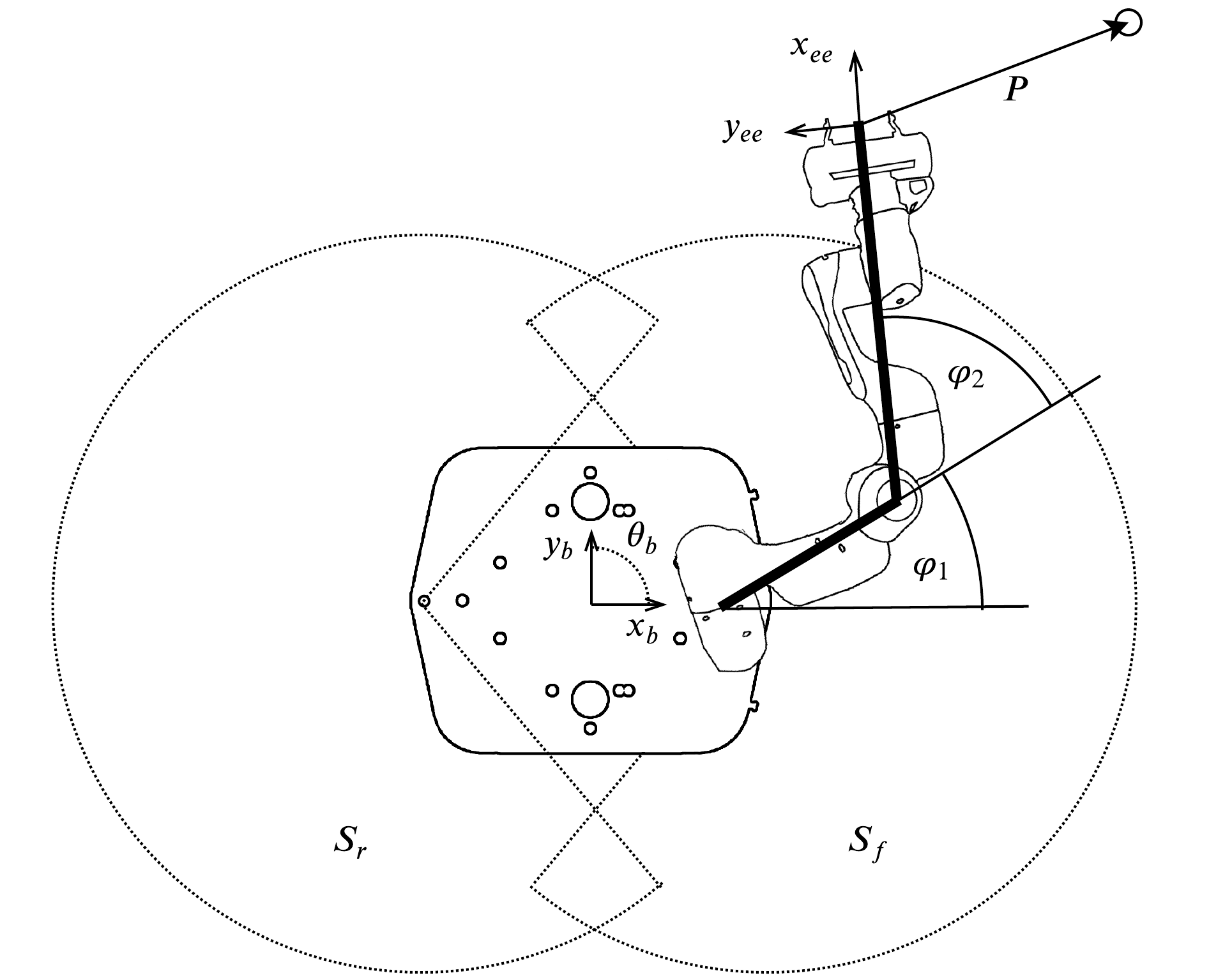}
  \caption{Top down view of our mobile manipulator RoyalPanda. We lock the arm configuration such that the arm can only move in a $2$D plane parallel to the floor with two actuated joints.}
  \label{fig:obsactspace}
\end{figure}

We model our mobile manipulator as shown in Fig.~\ref{fig:obsactspace}. The observation of the complete system consists of two \acrshort{lidar} scans ($\boldsymbol{S_f}$, $\boldsymbol{S_r}$), the joint positions ($\boldsymbol{\varphi}$), joint and base velocities ($\dot{\boldsymbol{\varphi}}$, $\dot{x}_b$, $\dot{y}_b$, $\dot{\theta}_b$)
and the setpoint position in end-effector frame ($\boldsymbol{P}$).
The actions are the (discretized) accelerations of the base ($\Ddot{x}_b$, $\Ddot{y}_b$, $\Ddot{\theta}_b$) and the joints ($\Ddot{\boldsymbol{\varphi}}$).
The action discretization factor and the kinematic and dynamic limits used are listed in Table~\ref{tab:params}.

\subsection{Simulation}\label{sec:method_simulation}
Due to the large amounts of data needed to train a (deep) neural network, an agent is first trained in simulation. For that, we created a 3D simulation of our mobile manipulator RoyalPanda (Fig.~\ref{fig:teaser}) using PyBullet~\cite{coumans_2019} while offering an OpenAI Gym interface~\cite{brockman2016openai}. 

In order to train an agent that is able to control the system in a variety of environments, we randomize the simulation at each episode. In this work, we focus on scenarios where a mobile manipulator needs to operate in a corridor-like environment. We prevent overfitting to a specific setup by varying the corridor width, number and location of shelves, walls and doors and their respective dimensions. Similarly, special care was taken to ensure that the trained agent will work in a realistic environment with imperfect sensors. We added Gaussian noise to the \acrshort{lidar} scans and randomly set measurements to the maximum distance (\SI{5}{\metre}).

\subsection{The agent}\label{sec:method_agent}
We have chosen to use \acrfull{ppo}~\cite{schulman2017proximal}, a policy-based \acrshort{rl} algorithm which is often used in the field of robotics~\cite{heess2017emergence, schulman2017proximal, mahmood2018benchmarking, tan2018sim,chen2019rocket,lopes2018ppoquad} due to the ease of hyperparameter-tuning while retaining the stability and reliability of \acrfull{trpo}~\cite{schulman2015trust}. \acrshort{ppo} is an actor-critic method which ensures that the policy does not change dramatically between batch updates, resulting in smoother, more stable and more robust learning behavior.

\begin{figure}
  \centering
  \vspace{4pt}
  \includegraphics[width=1\columnwidth]{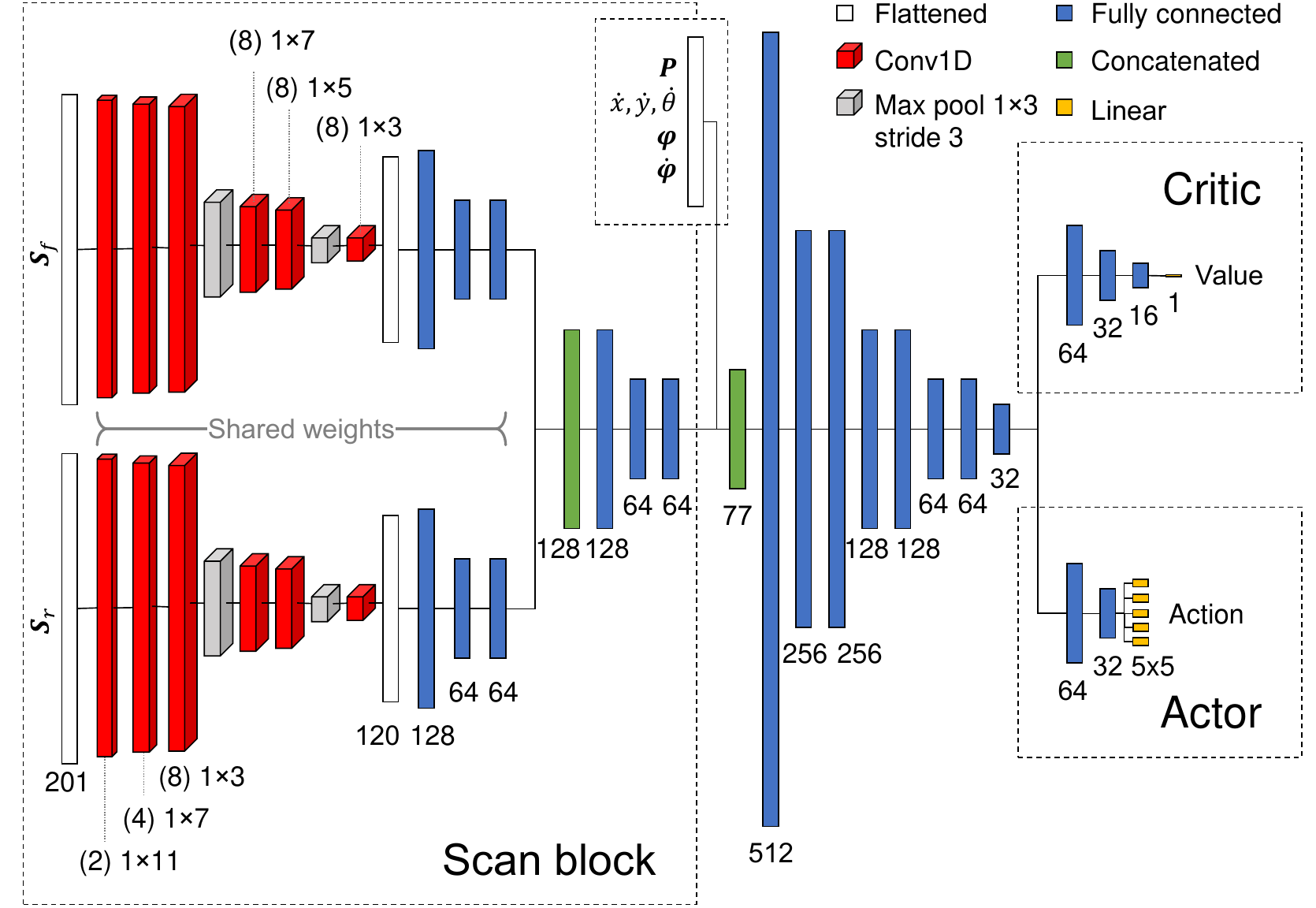}
  \caption{The network architecture used in our experiments. All activations are of type LeakyReLU, except for the last layers of the actor and critic which are linear.}
  \label{fig:network}
\end{figure}

The architecture used to train the agent is depicted in Fig.~\ref{fig:network}. Data from the two \acrshort{lidar} scans are compressed in the same branch composed of convolutional, max pooling and fully connected layers to $64$ floats. These are then combined and compressed to $64$ floats by three fully connected layers. At this point, the setpoint pose $\mathbf{P}$, base velocities $\dot{x}_b,\dot{y}_b,\dot{\theta}_b$, joint positions $\mathbf{\varphi}$ and joint velocities $\dot{\mathbf{\varphi}}$ are concatenated and further compressed by eight fully connected layers to $32$ floats. The model then separates into an actor and a critic, which result in the policy ($5\times5$ floats) and the value function ($1$ float), respectively. The actions are then chosen by evaluating the $\mathrm{argmax}$ of each block of $5$ floats.


\subsection{Training procedure}\label{sec:training_proc}
\acrshort{ppo} is an actor-critic method and can therefore be trained on data generated from multiple workers that run in parallel. We used Stable Baselines~\cite{stable-baselines}, a collection of \acrshort{rl} algorithms that offers an implementation called PPO2 with multiple workers as well as GPU and Tensorboard support.

Training was conducted with $32$ workers in parallel on an Intel Xeon E5-2640 v3 (8 Core @ \SI{2.6}{\giga\Hz}) and the network was updated on an Nvidia GeForce GTX Titan X. The agent was trained with $60$\,M steps in simulation which took about \SI{20}{\hour}. The learning parameters used in our experiments are listed in Table~\ref{tab:params}.

In order to speed up the training process and robustly achieve convergence, we used a simplified variant of \acrfull{adr}~\cite{akkaya2019rubixadr} on the size of the tolerance sphere around the setpoint ($d_h$).

\begin{table}[b]
 \centering
 \caption{Learning and reward function parameters}
 \vspace{-4pt}
 \begin{tabular}{ |p{1.7cm}|p{1.3cm}||p{1.7cm}|p{1.3cm}|  }
  \hline
  \multicolumn{4}{|c|}{Learning parameters} \\
  \hline
  noptepochs    & $30$          & cliprange     & $0.2$     \\
  cliprange\_vf & $-1$ (off)    & ent\_coeff    & $0.00376$ \\
  gamma         & $0.999$        & lam           & $0.8$     \\
  n\_steps      & $2048$        & nminibatches  & $8$       \\
  Learning rate & \multicolumn{3}{l|}{Linear drop from $\expnumber{1}{-3}$ to $\expnumber{0.15}{-3}$} \\
  \hline\hline
  \multicolumn{4}{|c|}{Reward function parameters} \\
  \hline
  $w_{t}$      & $-15$             & $T_t$      & $120$         \\
  $w_{pd}$         & $-10$             & $w_{pt}$               & $30$           \\
  $w_{sm}$      & $-1$             & $d_{th}$      & $0.3$         \\

  $w_{ht}$      & $20$             & $w_{hd}$      & $40$         \\
  $T_h$         & $1.5$   & $\tau$ & $0.04$      \\
  $D_c$         & $-60$             & $D_l$         & $-20$         \\
  $D_h$         & $10$             &               &           \\
  \hline\hline
  \multicolumn{4}{|c|}{Observation and action space limits}\\
  \hline
  $\overline{{S}}$ & \SI{5}{\metre} & $n_\text{action\_discr}$ & $5$ \\ $\overline{\abs{\dot{\varphi}}}$ & \SI{0.5}{\radian\per\second}   & $\overline{\abs{\Ddot{\varphi}}}$ & \SI{0.8}{\radian\per\square\second}  \\  $\overline{\abs{\dot{x}_b}},\overline{\abs{\dot{y}_b}}$   & \SI{0.1}{\metre\per\second} & $\overline{\abs{\Ddot{x}_b}},\overline{\abs{\Ddot{y}_b}}$& \SI{0.15}{\metre\per\square\second}\\
  $\overline{\abssmall{\dot{\theta}_b}}$   & \SI{0.2}{\radian\per\second} & $\overline{\abssmall{\Ddot{\theta}_b}}$& \SI{0.3}{\radian\per\square\second}\\

  \hline
 \end{tabular}
 \label{tab:params}
\end{table}

\subsection{Reward function}\label{sec:reward}

We use a handcrafted reward function:

\vspace{-10pt}
\begin{flalign} \label{eq:reward}
&r(\Delta d_{pd}, \Delta d_{pt}, \vec{v}, d_{sm}, d_g, d_h, \vec{D}, I_h) =&&\nonumber\\
&\,\,w_t \frac{\tau}{T_t} 
+ w_{pd} \Delta d_{pd} + w_{pt} \frac{\Delta d_{pt}}{d_{pt,init}}
+ w_{sm} \norm{\vec{v}} \tau L(d_{sm}, d_{th})&&\nonumber\\
&\,\,+ (w_{ht} + w_{hd} L(d_{g}, d_{h})) \frac{\tau}{T_{h}}
+ D_c + D_l + D_h - I_h,
\end{flalign}
where $L(x, y) = 1-\mathrm{min}(1, x/y)$. Individual components of $r$ are described in the following subsections and the weights used in our experiments are listed in Table~\ref{tab:params}.

\subsubsection{Time penalty, $w_{t}$}
We introduce a timeout $T_t$ after which the episode is terminated. We penalize time in each step to favour actions that lead to reaching the setpoint faster, summing up to a maximum (negative) value of $w_t$.

\subsubsection{Goal distance, $w_{pd}$, $w_{pt}$}
\begin{figure}
  \centering
  \vspace{4pt}
  \includegraphics[width=1\columnwidth]{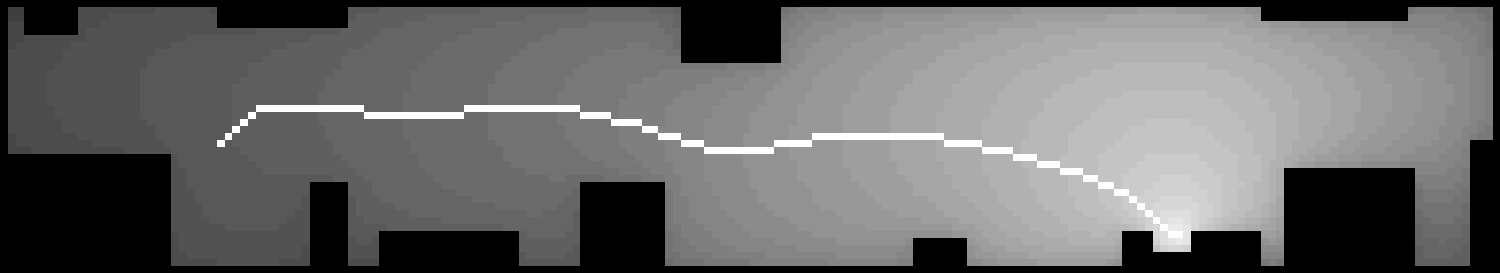}
  \caption{A \acrlong{hpt} which was used to generate a collision-free path for the end-effector to the goal position.}
  \label{fig:hpt}
\end{figure}

We use a \acrfull{hpt} to generate a collision-free path for the end-effector to the goal (Fig.~\ref{fig:hpt}) at the beginning of each episode. We then penalize deviation from the path ($\Delta d_{pd}$) by $w_{pd}$ and reward progress made along the path ($\Delta d_{pt}$) such that the maximum total reward is $w_{pt}$. Compared to a simple metric based on Euclidean distance to goal, our reward formulation allows the agent to easily learn to drive around obstacles and avoid getting stuck behind them in a local optima.

\subsubsection{Safety margin, $w_{sm}$}\label{sec:safetymargin}
We enforce a safety margin from obstacles for the base, by adding a term that linearly reduces the reward per driven distance with decreasing smallest distance to obstacle $d_{sm}$, starting with value $0$ at threshold distance $d_{th}$ and maximal reduction $w_{sm}$ at distance $0$.

\subsubsection{Holding goal position, $w_{ht}$, $w_{hd}$ and $I_h$}
Many tasks such as grasping an object require the end-effector to keep still. Thus, instead of ending an episode as soon as the end-effector reaches the goal, we require it to be within a tolerance sphere of radius $d_h$ (set by \acrshort{adr}, Section~\ref{sec:training_proc}) for a sustained interval $T_h$. The reward is incremented each time step the end-effector remains in the sphere, summing up to a total of $w_{ht}$. Additionally, we desire the goal distance $d_g$ to be as small as possible, therefore rewarding a smaller distance with a maximum total reward of $w_{hd}$ in a similar fashion as for the safety margin. To prevent the agent from exploiting the situation by repeatedly entering and leaving the sphere, we store the accumulated holding reward in $I_h$ and subtract it immediately if the agent leaves the sphere.

\subsubsection{Episode termination, $D_c$, $D_l$ and $D_h$}
Collision with an obstacle is penalized by $D_c$ and reaching the joint limits is penalized with $D_l$. Otherwise, the agent is rewarded with $D_h$ as soon as it achieves a sustained holding time of $T_h$. All three cases result in the termination of the current episode.

\subsection{Sampling-based motion planning baseline}
\label{sec:sampling_based_baseline}
We compared our implementation to a baseline sampling-based algorithm. In particular, we used the implementation of RRTConnect~\cite{Kuffner2000RRTC} in the MoveIt! motion planning framework. Since MoveIt! does not natively support planning movements of a mobile base platform, we introduced two prismatic and one continuous joint to model the movement of a holonomic platform in a plane. The trajectory received by the planner can then be followed with a simple controller. However, an accurate localization system is required for tracking the mobile base trajectory.

Due to the fact that MoveIt! is an offline motion planning library, we generated an octomap of our environment by using a Gazebo plugin from RotorS~\cite{Furrer2016RotorS}. This technique allows us to plan the trajectory to a setpoint in a single run but comes with the drawback that all obstacle positions must be static and known a priori. Further, to ensure reliable collision avoidance for the prismatic joints modelling the base platform, we reduced the search resolution of the kinematics solver to $1e-5$. We set the number of planning attempts to $20$ and limited the planning time to \SI{180}{\s}.



\section{Results}\label{sec:results}
\begin{figure}[ht]
  \includegraphics[width=1\columnwidth]{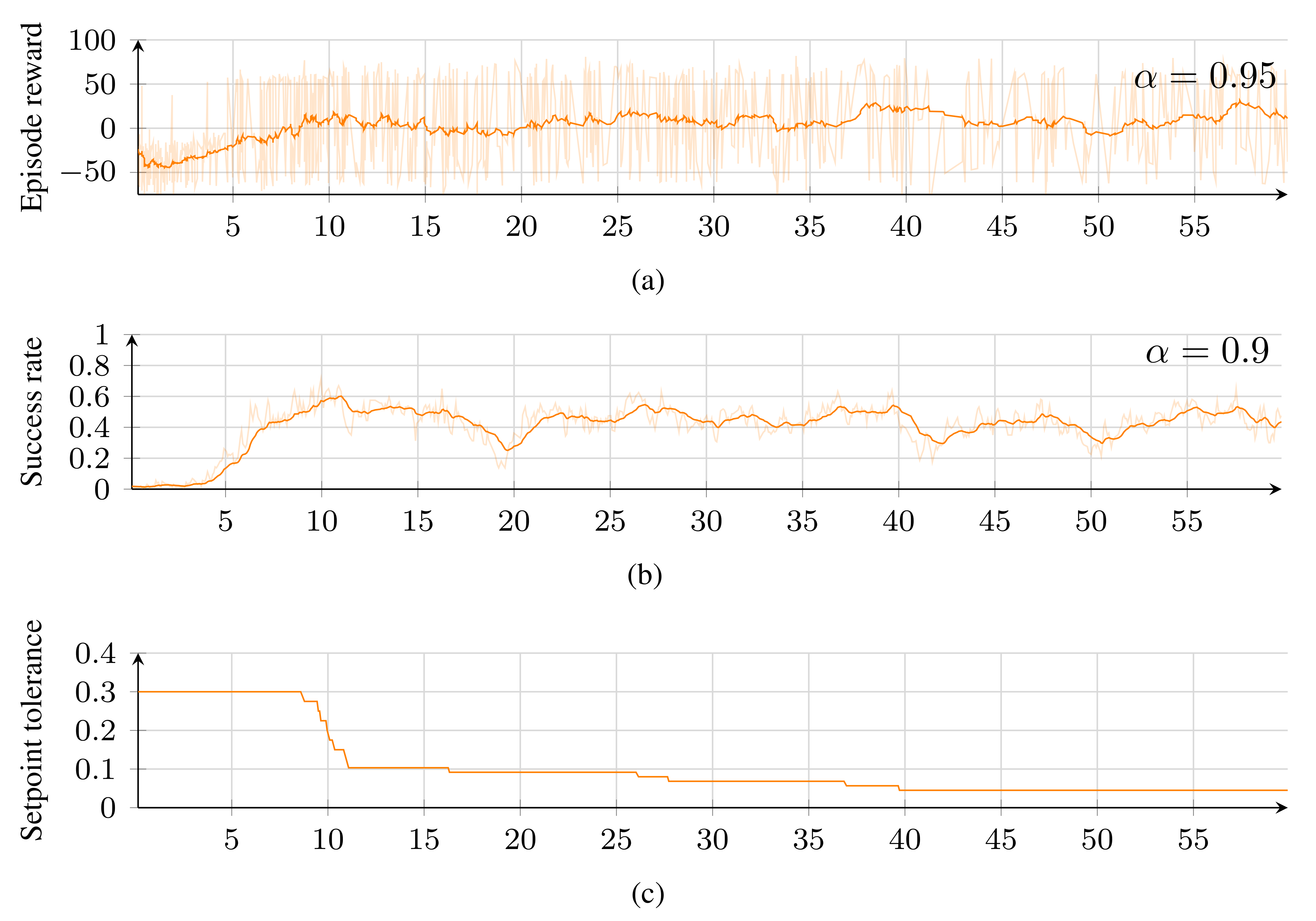}

    \caption{Accumulated reward per episode and success rate with their respective smoothed plots (exponential moving average) and setpoint tolerance during training.}
    \label{fig:training_plots}
\end{figure}

Figure~\ref{fig:training_plots} shows the accumulated reward per episode, the success rate and the \acrshort{adr} values during training.
We conducted our experiments on a Raspberry Pi Model 3B~\cite{raspi} with an ARM Cortex-A53 (4 Core @ \SI{1.2}{\giga\Hz}) and 1GB of \acrshort{ram}. Except for active cooling, we did not add any equipment (such as a Neural Compute Stick) to the device and therefore run the network directly on the CPU which required about $160$MB of \acrshort{ram}. We achieved a network prediction rate of \SI{104}{\Hz}.

The trained agent was first evaluated in a Gazebo simulation (run on a separate machine) which modelled various corridor environments before we deployed it to the real system. We compared the performance against the RRTConnect planning baseline (Section~\ref{sec:sampling_based_baseline}). The metrics used for comparing the two approaches are the total mission time, as a sum of the planning and execution time, and the total travelled base and joint distances. We additionally highlight the planning and execution times and the success rate for the two methods. Note that failure for the \acrshort{rl} agent occurs due to collision or execution timeout (\SI{180}{\s}), while RRTConnect failure occurs when the planner is unable to find a collision-free path in the given time. The setpoint tolerance was fixed to a value of \SI{0.07}{\metre}.

\subsection{Simulation results}
\begin{figure}[t]
  \centering
  \vspace{4pt}
  \begin{tikzpicture}
    \node[anchor=south west,inner sep=0] (image) at (0,0) {\includegraphics[width=1\columnwidth]{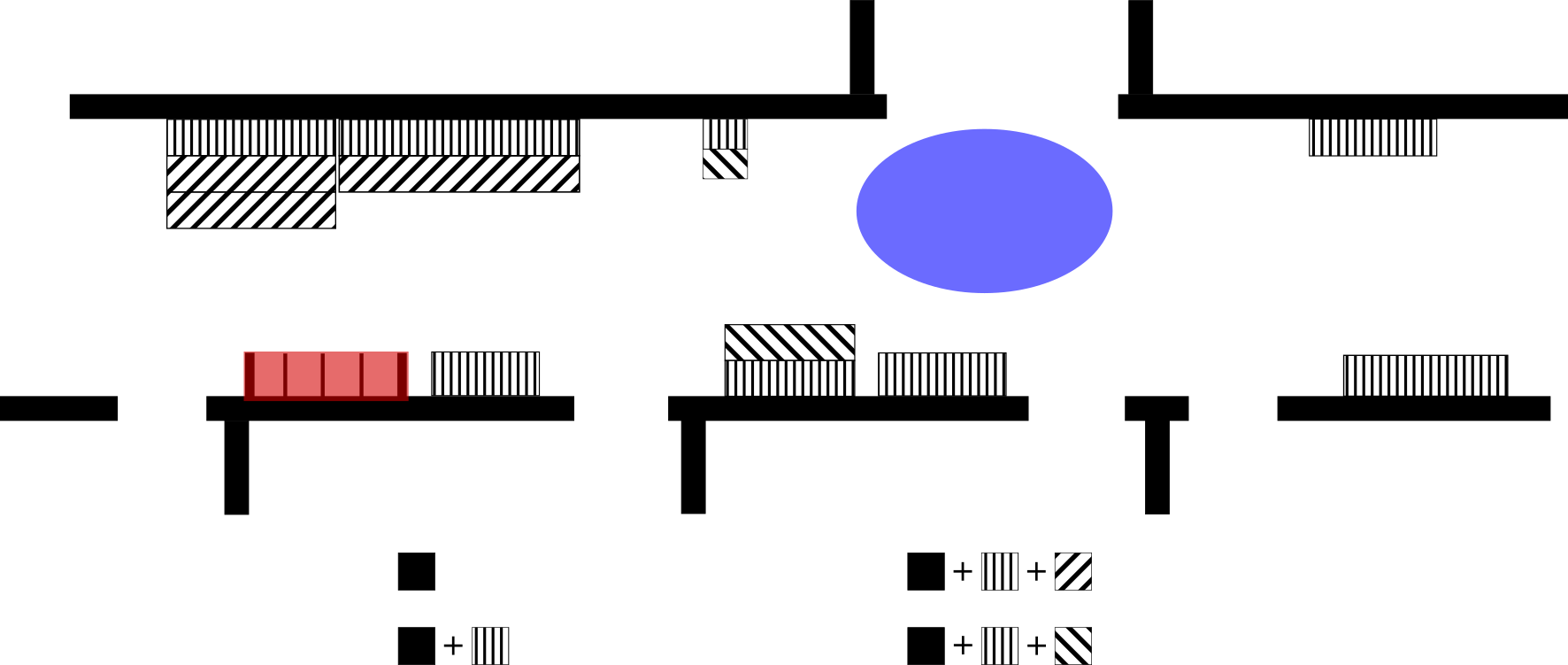}};
    \begin{scope}[x={(image.south east)},y={(image.north west)}]
    
        \node[left] (t1) at (0.25, 0.149) {\small Task 1: };
        \node[left] (t2) at (0.25, 0.032) {\small Task 2: };
        \node[left] (t3) at (0.575, 0.149) {\small Task 3: };
        \node[left] (t4) at (0.575, 0.032) {\small Task 4: };
        
        
       
    \end{scope}
\end{tikzpicture}
  
  \caption{Collision map of the simulated environments for tasks 1-4 used in our evaluation. The red square denotes the shelf in which the setpoints are spawned, the blue ellipse marks the region of initial poses.}
  \label{fig:simutasks}
\end{figure}
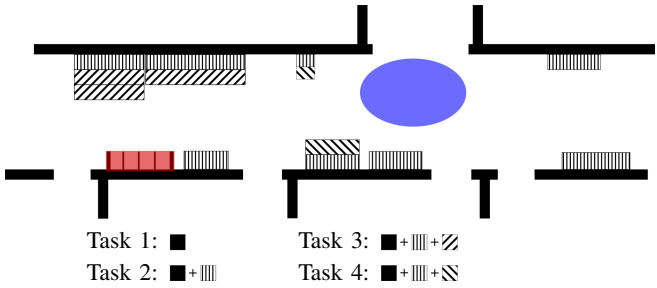

Figure~\ref{fig:simutasks} shows four different scenarios in which we evaluated the \acrshort{rl} agent and RRTConnect. The four scenarios vary in difficulty according to the placement of obstacles in the environment. We randomly sampled a feasible initial configuration inside the blue ellipse and a setpoint in the red shelf and ran both algorithms. Table~\ref{tab:simresults} shows the averaged metrics from 100 runs for each task.

Even though RRTConnect achieved a lower execution time in all tasks, due to the long planning time required, our approach had a consistently lower total mission time. In Task 2, our approach not only achieved a lower total mission time but also a higher success rate. In the remaining tasks, however, the success rate was notably lower. This might come from the fact that our procedure for environment randomization during training was more likely to produce scenes similar to Task 2 compared to the other tasks.

In all tasks, RRTConnect achieved smaller base and joint distances compared to the \acrshort{rl} agent. The learned agent tended to exhibit small oscillatory behaviors whose presence may be due simply to the fact that we did not penalize base and joint distances in our reward function. 

Note that even though the evaluations were done on static scenes, unlike RRTConnect, our approach can also handle dynamic scenes since we are controlling the system in a closed-loop fashion.



\begin{table*}[bt]
 \centering
 \vspace{8pt}
 \caption{Average simulation results of the \acrshort{rl} agent and RRTConnect over 100 runs. Standard deviations are provided in brackets.}
 \vspace{-4pt}
 \begin{tabular}{ |c|l|l|l|l|l|l|l|  }
  \hline
  \multirow{2}{*}{Task} & \multirow{2}{*}{Method} & Total mission time & Base distance & Joint distance & Planning time & Execution time & Success\\
  & & [s] & [m] & [rad] & [s] & [s] & rate\\
  \hline
  
  \multirow{2}{*}{1}           & \acrshort{rl} agent    & {$\mathbf{77.26\,(9.82)}$}    & $6.71\,(0.71)$      & $16.88$ ($3.56$)     & - & $77.26\,(9.82)$ & $0.55$     \\
  & RRTConnect & $82.07\,(20.93)$    & $\mathbf{5.93\,(1.61)}$  & $\mathbf{2.72\,(1.35)}$     & $22.83\,(11.75)$ & $\mathbf{59.24\,(16.15)}$ & $\mathbf{0.85}$     \\
 
  \hline
  \multirow{2}{*}{2}           & \acrshort{rl} agent    & $\mathbf{79.31\,(14.84)}$    & ${6.63\,(1.02)}$      & $18.71\,(4.24)$     & - & ${79.31\,(14.84)}$ & $\mathbf{0.79}$     \\
                                    & RRTConnect &$ 89.73\,( 14.82 )$     & $ \mathbf{5.96\,( 0.76 )}$      & $ \mathbf{3.07\,( 1.45 )}$     & $ 30.13\,( 12.85 )$ & $ \mathbf{59.61\,( 7.69 )}$ & ${0.77}$     \\
  \hline

  \multirow{2}{*}{3}           & \acrshort{rl} agent    & $\mathbf{76.7\,(13.74)}$    & ${6.4\,(0.93)}$      & $16.98\,(3.66)$     & - & ${76.7\,(13.74)}$ & $0.56$     \\
                                    & RRTConnect & $ 102.05\,( 39.95 )$    & $ \mathbf{5.75\,( 1.06 )}$       & $ \mathbf{3.33\,( 1.66 )}$      & $ 44.77\,( 39.32 )$  & $ \mathbf{57.28\,( 10.66 )}$  & $\mathbf{0.66}$     \\
  \hline
 
  \multirow{2}{*}{4}           & \acrshort{rl} agent   & $\mathbf{81.45\,(16.66)}$    & $\mathbf{6.86\,(1.0)}$      & $19.09\,(3.94)$     & - & ${81.45\,(16.66)}$ & $0.61$     \\
                                    & RRTConnect & $ 104.86\,( 42.84 )$    & $ 7.17\,( 4.0 )$      & $ \mathbf{3.94\,( 1.84 )}$     & $ 33.24\,( 13.4 )$ & $ \mathbf{71.62\,( 38.43 )}$ & $\mathbf{0.85}$     \\
  \hline
  
  
  \multirow{2}{*}{Avg.}           & \acrshort{rl} agent   & $\mathbf{78.8\,(14.26)}$    & ${6.65\,(0.95)}$      & $18.02\,(4.02)$     & - & ${78.8\,(14.26)}$ & $0.63$     \\
                                    & RRTConnect & $ 94.59\,( 33.4 )$    & $ \mathbf{6.26\,(2.45 )}$      & $ \mathbf{3.28\,( 1.66 )}$     & $ 32.1\,( 22.45 )$ & $ \mathbf{62.48\,( 23.79 )}$ & $\mathbf{0.80}$     \\
  \hline
 \end{tabular}
 \label{tab:simresults}
 \vspace{-0.5cm}
\end{table*}

\subsection{Real-world experiments}
We verified the trained \acrshort{rl} agent through $20$ experimental runs on our mobile manipulator RoyalPanda. Maplab~\cite{schneider2018maplab} was used for the estimation of the setpoint pose $\mathbf{P}$ in the end-effector frame. The agent ran on a laptop with an Intel i7-8550U (4 Core @ 1.8GHz) with 16GB of \acrshort{ram}. We deployed the robot in a corridor of our lab and tested the limitations of the agent with different initial robot configurations and a cardboard box to change the corridor width in different locations (see Fig.~\ref{fig:teaser}). 

The experiments showed that the agent\----which was solely trained in simulation\----can be transferred to a real system. For a normal corridor environment similar to Task 2 (Fig.~\ref{fig:simutasks}), the agent was consistently reaching the setpoint. However, for more difficult scenarios such as a tight corridor, the base platform sometimes collided into our cardboard box. A possible explanation is that the robot has two blind spots which are invisible to the \acrshort{lidar} sensors. As soon as an obstacle corner is inside these, the agent seemed to cut the corner in order to reach the goal as quickly as possible, thus colliding with the box. This issue could be addressed by adding a memory structure to the network to help retain information from previous scans.
Also, depending on the angle of the end-effector to the setpoint, the manipulator sometimes collided with the shelf. This may be due in part to an inaccurate manipulator-to-base calibration.

The accompanying video, which was also uploaded to Youtube\footnote{\url{https://youtu.be/3qobNCMUMV4}}, shows some of the real-world experiments.

\section{Conclusion} 
\label{sec:conclusion}
In this work, we have shown a method to learn \acrshort{wbc} of a mobile manipulator by training a (deep) neural network with \acrshort{rl}. We used \acrshort{adr} to automatically increase the complexity of the problem with growing success rate to speed up training and achieve robust convergence. We compared our agent against RRTConnect on an embedded device in simulation and showed that our approach achieves consistently shorter total mission times.
Furthermore, we showed that the trained agent can be directly transferred to a real robot.

In future work, the success rate of our approach could be increased by adding a safety controller which separately limits the action space for the base platform and the manipulator to avoid collisions. Furthermore, the used \acrshort{dof} of the manipulator could be increased by adding a depth sensor as observation, potentially in an encoded form generated by an Auto-Encoder network.


\bibliographystyle{IEEEtranN}
\footnotesize
\bibliography{IEEEabrv,references}

\end{document}